\documentclass{article}

\usepackage[preprint]{nips_2018}

\usepackage[utf8]{inputenc} % allow utf-8 input
\usepackage[T1]{fontenc}    % use 8-bit T1 fonts
\usepackage{hyperref}       % hyperlinks
\usepackage{url}            % simple URL typesetting
\usepackage{booktabs}       % professional-quality tables
\usepackage{amsfonts}       % blackboard math symbols
\usepackage{nicefrac}       % compact symbols for 1/2, etc.
\usepackage{microtype}      % microtypography

\usepackage{amsmath}
\usepackage{amssymb}
\newcommand{\sgn}{\mathop{\mathrm{sgn}}}

\usepackage{float}
\usepackage{graphicx}
\usepackage{rotating}
\graphicspath{{./}{../png/}{../eps/}{../asy/}}
\newcommand{\defwidth}{\textwidth}
\newcommand{\graphic}[2][width=\defwidth]{\centerline{\includegraphics[#1]{#2}}}
\newcommand{\optlabel}[1]{\ifx&#1& \else \label{#1} \fi}
\newcommand{\legend}[2][]{\makebox[\defwidth][c]{\begin{minipage}[t]{1.\defwidth}\caption{\normalsize{#2}}\optlabel{#1}\end{minipage}}\\}
\newenvironment{portrait}[1][!htp]{\begin{figure}[#1] \begin{centering}}{\end{centering}\end{figure}}

\title{Deep learning improved by biological activation functions}

\author{
	Gardave S.~Bhumbra\thanks{Website: \url{www.deepnodal.net}} \\
  Department of Neuroscience, Physiology and Pharmacology,\\
  UCL, Gower St.,\\
  London (UK), WC1E 6BT.\\
  \texttt{g.bhumbra@ucl.ac.uk} \\
}

\begin{document}

\maketitle

\begin{abstract}
`Biologically inspired' activation functions, such as the logistic sigmoid, have been instrumental in the historical
advancement of machine learning. However in the field of deep learning, they have been largely displaced by rectified
linear units (ReLU) or similar functions, such as its exponential linear unit (ELU) variant, to mitigate the effects of
vanishing gradients associated with error back-propagation. The logistic sigmoid however does not represent the true
input-output relation in neuronal cells under physiological conditions. Here, bionodal root unit (BRU) activation
functions are introduced, exhibiting input-output non-linearities that are substantially more biologically plausible
since their functional form is based on known biophysical properties of neuronal cells.

In order to evaluate the learning performance of BRU activations, deep networks are constructed with identical
architectures except differing in their transfer functions (ReLU, ELU, and BRU). Multilayer perceptrons, stacked
auto-encoders, and convolutional networks are used to test supervised and unsupervised learning based on the MNIST and
CIFAR-10/100 datasets. Comparisons of learning performance, quantified using loss and error measurements, demonstrate
that bionodal networks both train faster than their ReLU and ELU counterparts and result in the best generalised models
even in the absence of formal regularisation. These results therefore suggest that revisiting the detailed properties of
biological neurones and their circuitry might prove invaluable in the field of deep learning for the future.
\end{abstract}

\sloppy

\section{Introduction}

Neurobiology has inspired the design of computational networks ever since they were first conceived
\citep{hassabis2017neuroscience}. When the `perceptron' classifier was introduced \citep{rosenblatt1958perceptron}, the
non-linearity used to characterise the input-output relationship was based on a critical input boundary above which
outputs were fully activated, resembling the `threshold' and `all-or-none' properties of neurones
\citep{hodgkin1952quantitative}. Continuous `activation' or `transfer' functions, such as the sigmoid non-linearity
(Figure \ref{fig:trf}), were employed subsequently to represent the output of units as analogous to neuronal firing
rates \citep{dayan2001theoretical}. 

With the advance of deep learning however `biologically-inspired' activation functions have been largely supplanted by
non-linearities that possess a linear component in the positive polarity \citep{nair2010rectified,glorot2011deep}. A key
advantage of these more recent transfer functions, notably the rectified linear unit (ReLU, Figure \ref{fig:trf}), is
that their non-contractive positive activations mitigate the effects of vanishing gradients associated with error
back-propagation. Since the deep learning performance of the ReLU activation function and its variants usually exceed
that of their sigmoid predecessors that confer no such advantage, the utility of `biological insights' into input-output
relationships might be called into question.

\begin{portrait}[h] %[P] 
	
	%\graphic[width=1.3\defwidth,natwidth=4125,natheight=1283]{trf.png} 
\graphic[width=1.\defwidth,trim={0cm 0 0cm 0cm}]{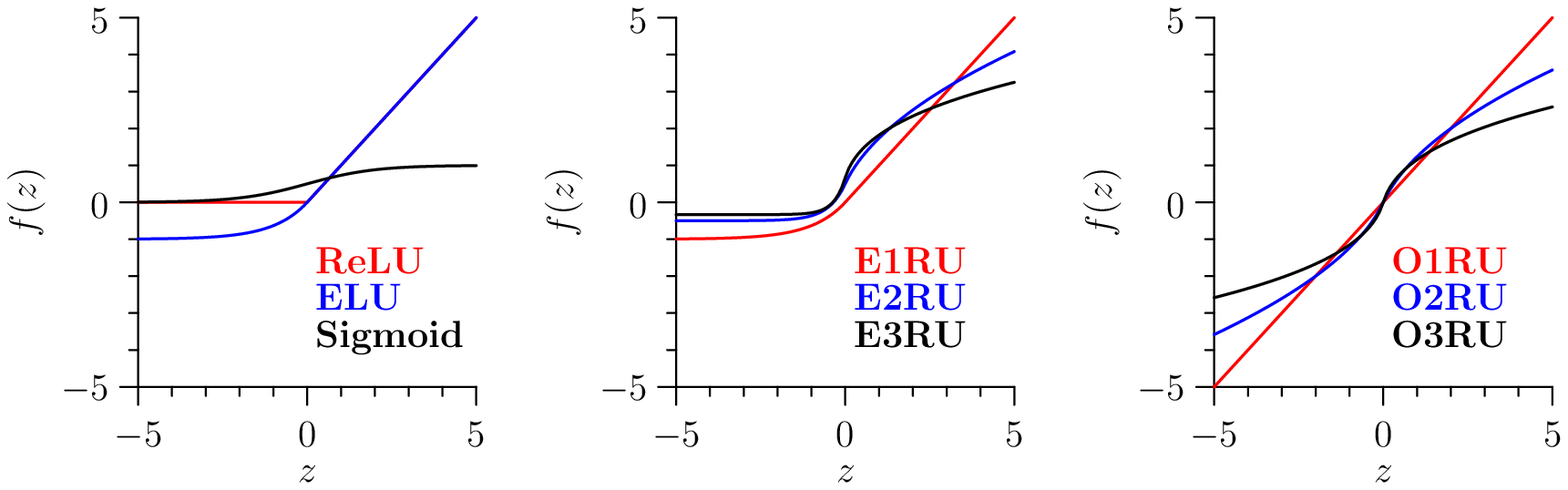} 

	\legend[fig:trf]{Graphs of activation functions. The left overlays the standard logistic sigmoid function, rectified
	linear unit (ReLU), and exponential linear unit (ELU). Bionodal root unit (BRU) transfer functions are shown for 
	exponential root units (ERU, middle) and odd root units (ORU, right), for radices: $r = \{1, 2, 3\}$. }
\end{portrait}

Electrophysiological recordings \citep{binder2011physiological} and biophysical simulations
\citep{fourcaud2003spike,bhumbra2014recurrent} of neurones however show that their input-output relationships are not
sigmoid. At rheobase, which defines the minimum positive current required to evoke a single output discharge, only
modest increments in depolarising inputs are required to accelerate firing in many neuronal types. This acceleration
results from active conductances, mediated by voltage-sensitive ion channels, that overcome passive leak conductances
and drive neuronal firing of action potentials. However, after an initial linear slope in the input-output relation, the
response curve only partially saturates as the voltage-sensitive channels become increasingly inactivated. 

It is possible to induce depolarisation experimentally to the extent of inactivating voltage-sensitive channels
completely. While this observation is consistent with the theoretical concept of a maximum output discharge rate, the
size of the necessary depolarisation is well beyond the physiological range. Therefore the suggestion that the
input-output relationship in biological neurones fully saturates represents as little useful scientific insight as the
suggestion that the ReLU activation must saturate in practice due to the finite numeric range of computer floating point
units.

Here, a family of transfer functions termed bionodal root units (BRU), are introduced which are substantially more
biologically plausible than sigmoid activations. BRU activation functions were devised on basis of the biophysical
properties of neuronal cells rather than as a result of theoretical optimisations of deep learning models. The present
study compares the deep learning performance of bionodal networks to that of identical networks that differ only in
their activation using either the ReLU transfer function or one of its variants \citep{clevert2015fast} called the
`exponential linear unit' (ELU, Figure \ref{fig:trf}).  

\section{Methods}

\subsection{Bionodal root units}

BRU activations comprise two subfamilies, namely the exponential root units (ERU) and odd root units (ORU) which can be
combined for different layers within a deep network. The exponential root unit (ERU) is:

\begin{align}
	f(z) =
  \begin{cases}
		(r^2 z+1)^{\frac{1}{r}} - \frac{1}{r}, & \text{if } z\geq 0\\
		e^{rz} - \frac{1}{r},                  & \text{if } z <   0
	\end{cases}
	&\phantom{,}&, &\phantom{MMM} 
	\frac{df(z)}{dz} =
  \begin{cases}
		r (r^2 z+1)^{\frac{1-r}{r}}, & \text{if } z\geq 0\\
		r e^{rz},                    & \text{if } z <   0 \label{eq:rru}
	\end{cases}
\end{align}

where the radix $r$ is a positive shape parameter that defines the choice of non-linearity. The ERU is monotonic and
its derivative is continuous everywhere including at $z=0$, where the derivative is at its maximum $r$. A root
function was chosen on the basis of the successful use of square root functions to fit current-frequency relationships
observed in current-clamp recordings from neurones \citep{ermentrout1998linearization} and those obtained from
biophysical models \citep{fourcaud2003spike}. Adoption of a square root function is analogous to an ERU with a radix $r$
of 2, which is termed the exponential 2nd root unit (E2RU, Figure \ref{fig:trf}). The first root (E1RU, Figure
\ref{fig:trf}) is equivalent to the ELU transfer function. For higher root units, the derivative is decreasing in the
positive polarity but does not asymptote to zero, unlike in the negative polarity. At the limit where $r\to \infty$, the
ERU tends to a Heaviside step function.

The odd radical root unit (ORU) is:

\begin{align}
	f(z) = \sgn(z) ( (r^2\lvert z \rvert + 1)^{\frac{1}{r}} - 1)
	&\phantom{,}&, &\phantom{MMM} 
	\frac{df(z)}{dz} = r (r^2 \lvert z \rvert + 1)^{\frac{1-r}{r}} \label{eq:oru}
\end{align}
	
where once again the radix $r$ is a positive shape parameter that defines the choice of non-linearity. Like the ERU,
the ORU is monotonic and its derivative is continuous everywhere including at $z=0$, where the derivative is at its
maximum $r$. However unlike the ERU, the symmetrical functional form of the ORU makes the activation function odd. The
first root (O1RU, Figure \ref{fig:trf}) corresponds to the identity function. For higher root units, the magnitude of
the derivative decreases in both polarities but does not asymptote to zero. Therefore while the profile of ORU
functions may resemble the hyperbolic tangent, ORU activations are not asymptotically bounded. 

Methods of initialising weight coefficients based on scaling second moments \citep{glorot2010understanding} aim to
regulate variances of activations propagated throughout hidden layers. Recent optimisation of this technique for ReLU
networks \citep{he2015delving} adopts a Gaussian initialisation with variances scaled by $2/n_i$, where $n_i$ is the
number of inputs. While such scaling would be suitable for BRU transfer functions with a radix $r$ of $1$, it is not
optimal for the greater curvatures exhibited by higher radices. The variance $\sigma^2$ scaling adopted for BRU layers
was therefore assigned empirically using reciprocal relations with the radix $r$:

\begin{align}
	\sigma^2 (\textrm{ERU}) &= \frac{6}{n_i(2r + 1)} \\
	\sigma^2 (\textrm{ORU}) &= \frac{2}{n_i r}
\end{align}

When $r$ is $1$, this would result in a scaling for E1RU and O1RU weight initialisation that would be identical to the
standard fan-in method \citep{he2015delving}. Output layers employing softmax or sigmoid functions were initialised
using identical variance scaling. During experiments, ReLU and ELU network weight coefficients were also initialised in
the same way. In all layers with weight coefficients, bias offset parameters were trainable and initialised to
$\mathbf{b} = \mathbf{0}$. No attempt was made to tune learning rates, which were set to a constant value at an order of
magnitude that resulted in no initial divergent learning in networks for any the three transfer functions. Experiments were run for 200 epochs, and conducted using the TensorFlow machine learning framework
\citep{tensorflow2015-whitepaper}.

\section{Results}

\subsection{Multilayer perceptron supervised learning}

Since the two subfamilies of activations functions as well as the corresponding weight initialisation procedures
presented in this study are novel, simple experiments were first conducted to confirm that BRU non-linearities
facilitate learning in networks of increasing depth. Fully connected multilayer perceptron networks, with the number of
hidden layers ranging from 4 to 8, were trained on the MNIST data set (60000 training and 10000 test samples of 28x28
images). In order to allow comparisons with previous work \citep{clevert2015fast}, the architectures and training
parameters were matched with 128 units in each hidden layer, a learning rate $\eta$ of 0.01, a batch size of 64, a
softmax output activation, a cross entropy loss function, and a stochastic gradient descent learning update. In
correspondence with the MNIST dataset dimensionality, the input data dimension was 784, and the number of output units
was 10.

Transfer functions for hidden layers were either ReLU, ELU, or BRU functions. BRU subtypes were assigned
according to the following scheme: O3RU for the first hidden layer, E2RU for the last hidden layer, and O2RU for
intervening hidden layers. This permutation was selected to facilitate back-propagation of gradients, to center
activations within the inner hidden layers close to zero, and to produce a sparse encoding input to the softmax output
layer. Initial training and test loss values are plotted for each the three types of networks in Figure
\ref{fig:mlpmnist}.

\begin{portrait}[!htp] %[P] 
	
	\graphic[width=1.\defwidth,trim={0 0 0 0},clip]{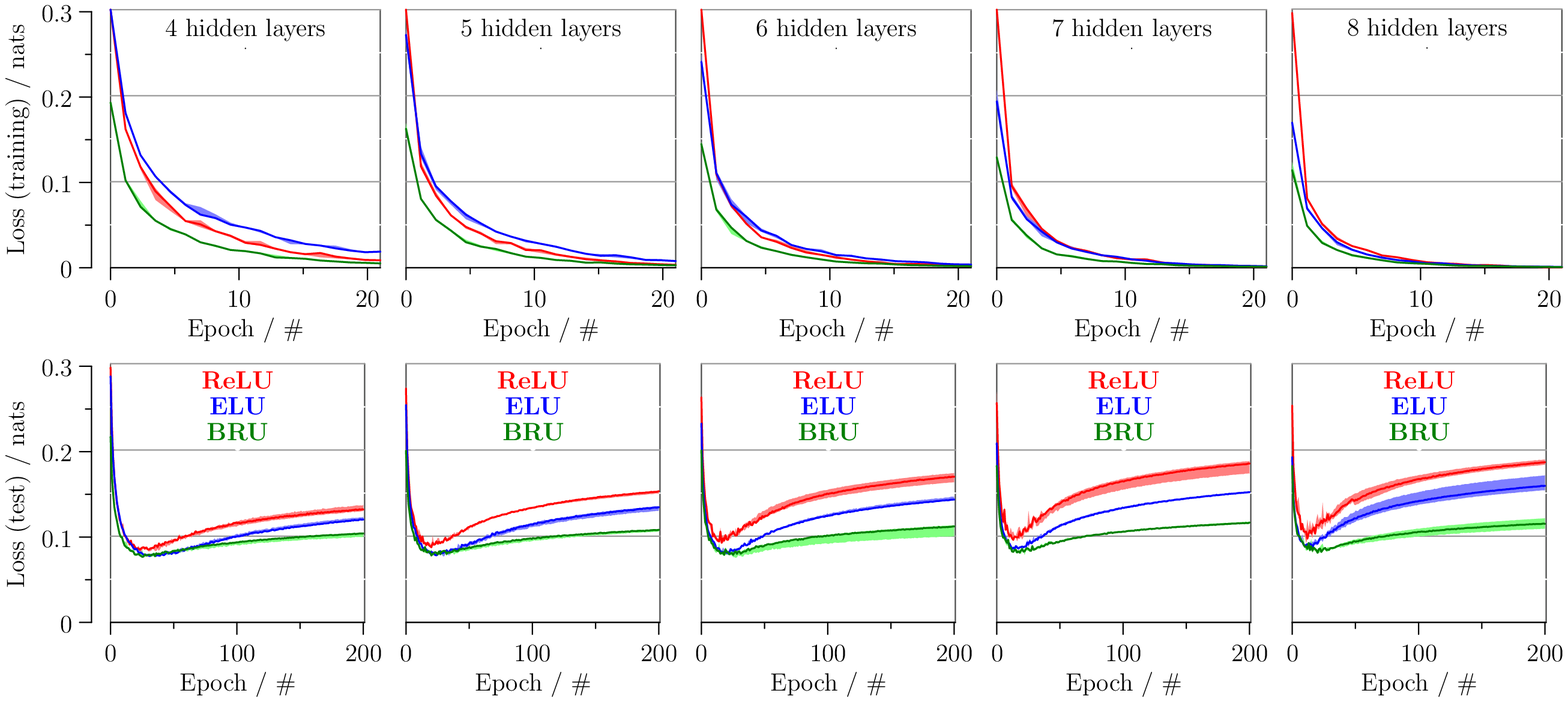} 

	\legend[fig:mlpmnist]{Multilayer perceptron training on the MNIST data set with 4 to 8 hidden layers. Networks
	employing ReLU, ELU, and BRU activation functions were trained, and the loss for training and test sets from several
runs are plotted as lines for the medians and bands for the inter-quartile ranges.}

\end{portrait}

Training cross entropy losses plotted in the top row of Figure \ref{fig:mlpmnist} show that with 4 hidden layers, the
ELU network is the slowest in training. While the ReLU network is somewhat faster, the bionodal network is the fastest.
With 6 hidden layers, the training performance of the ReLU and ELU networks have both improved with similar training
loss profiles but the RRU network remains faster. For the 8 hidden layer network, which repeats precisely the experiment
of \citet{clevert2015fast}, the ELU network outperforms the ReLU network, confirming the findings of the previous study.
The bionodal network however still maintains its lead in performance as the fastest learning network of the three.

Test cross entropy losses plotted in the bottom row of Figure \ref{fig:mlpmnist} show that with 4 hidden layers, ELU
networks generalise better ReLU networks. While the bionodal network shows the fastest reduction in test loss, it is
also relatively refractory to test loss increases compared to the other networks at the overfitting stage. As the number
of hidden layers is increased, the ReLU and ELU networks increasingly overfit with the ReLU networks being the more
sensitive. By contrast, overfitting in the bionodal networks is less affected by changes in the number of hidden layers,
and results in the lowest minimum test loss of the three networks during training with 8 hidden layers. This suggests
that the intrinsic capability of bionodal networks to learn generalised models is resistant to the adverse effects of
overfitting resulting from increases in depth within the network even in the absence of explicit regularisation.

\subsection{Stacked auto-encoder unsupervised learning}

Having established BRU transfer functions facilitate supervised learning in networks of increasing depth, a similar
experiment was performed to assess unsupervised learning using a stacked auto-encoder network. A standard auto-encoder
network design \citep{hinton2006reducing,desjardins2015natural,clevert2015fast} was implemented using a symmetrical
architecture with [1000, 500, 250, 30, 250, 500, 1000] units that include both encoder and decoder components with 784
inputs and 784 output units for the MNIST image data. Training parameters were kept constant using a batch size of 120,
a sigmoid transfer function for the output layer, a cross entropy loss function, an Adam \citep{kingma2014adam} learning
update ($\eta = 0.0001$, $\beta_1 = 0.9, \beta_2 = 0.999, \epsilon=0.001$), and weights for the encoder and decoder
components were untied. 

Transfer functions for hidden layers were either ReLU, ELU, or BRU functions. For BRU networks, the O2RU transfer
function was used for all hidden layers apart for the last which was assigned an E1RU function prior to the output
layer. In Figure \ref{fig:saemnist}, the loss and reconstruction errors are plotted for each the three types of networks
for the training test data. The results confirm the findings of \citet{clevert2015fast}, who show that ELU networks
show improved performance over ReLU networks for this auto-encoder architecture. However, bionodal networks outperform
both ELU and ReLU networks in all measures of performance.

\begin{portrait}[!htp] %[P] 
	
	\graphic[width=1.\defwidth,trim={0 0 0 0},clip]{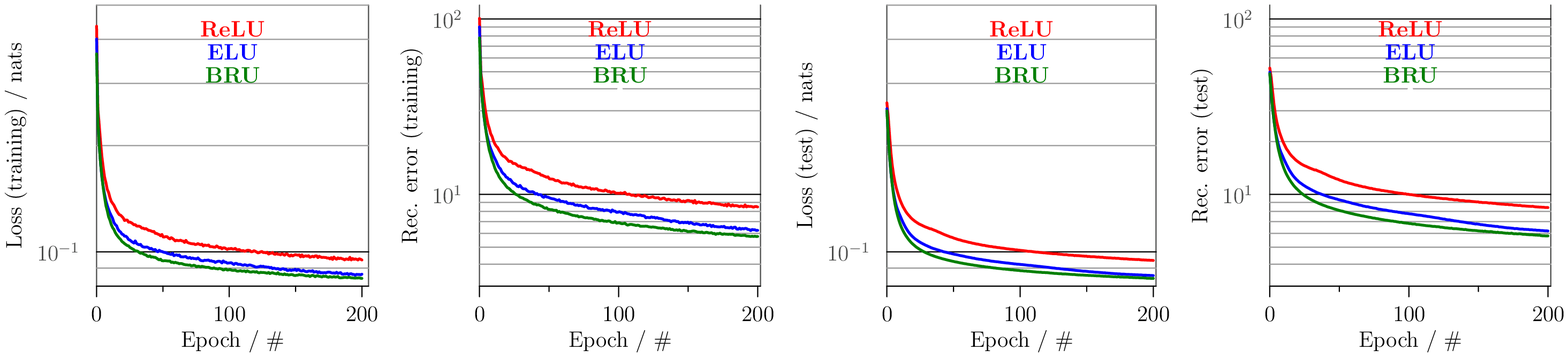} 

	\legend[fig:saemnist]{Stack auto-encoder training on MNIST data. Networks employing ReLU, ELU, and BRU activation
	functions were trained, and the loss and reconstruction errors for training and test sets from several runs are
  plotted as lines that represent medians.}

\end{portrait}

\subsection{Convolutional networks with MNIST data}

Since convolutional networks are well adapted for image recognition, they constitute architectures that are suited for
the MNIST data set. A convolutional network with a simple architecture was employed to moderate computational complexity
for fair and efficient comparison in performance between the activation functions. The architecture was based on the
LeNet-5 network \citep{lecun1998gradient}: [6 feature maps, 5x5 convolution kernel size, 1 stride], [6 maps, 2x2
average-pool, 2 stride], [16 maps, 5x5 convolution, 1 stride], [16 maps, 2x2 average-pool, 2 stride], [120 maps, 5x5
convolution, 1 stride], 84 fully-connected. 

The convolutional network was simplified from the LeNet design in three respects. First, `same' padding was used
throughout all convolution and pooling layers. Second, while average pooling layers were used, they did not include
learnable parameters and were assigned activation non-linearities for all networks. Finally, output activations were
softmax functions with losses calculated using cross entropies. Training and optimisation parameters were identical to
those used for the stack auto-encoder experiments described above. Images were zero-centered and rescaled to unit
variance.  BRU transfer function assignments were as follows: E1RU for all pooling layers, and [E3RU, O2RU, O2RU, E2RU]
for the convolution and hidden fully-connected layers.

In Figure \ref{fig:lenetmnist},  the loss and error quotients are plotted for each of the three types of networks for
the training data during the initial stage of learning and for the test data for all epochs. The training results
support the findings of \citet{clevert2015fast}, who show that ELU convolutional networks train faster than ReLU
networks. The bionodal network however exhibits the fastest learning. While the ELU network initially trains faster than
the ReLU network, the test loss and errors quotients show that learning in the ReLU network results in better
generalisation. However, bionodal networks outperform both ELU and ReLU networks as measured by the minimum test loss
and minimum test error quotient during training. These results suggest that even in the absence of explicit
regularisation, the intrinsic capability of bionodal networks to learn generalised models exceeds that of the other
networks. 

\begin{portrait}[!htp] %[P] 
	
	\graphic[width=1.0\defwidth,trim={0 0 0 0},clip]{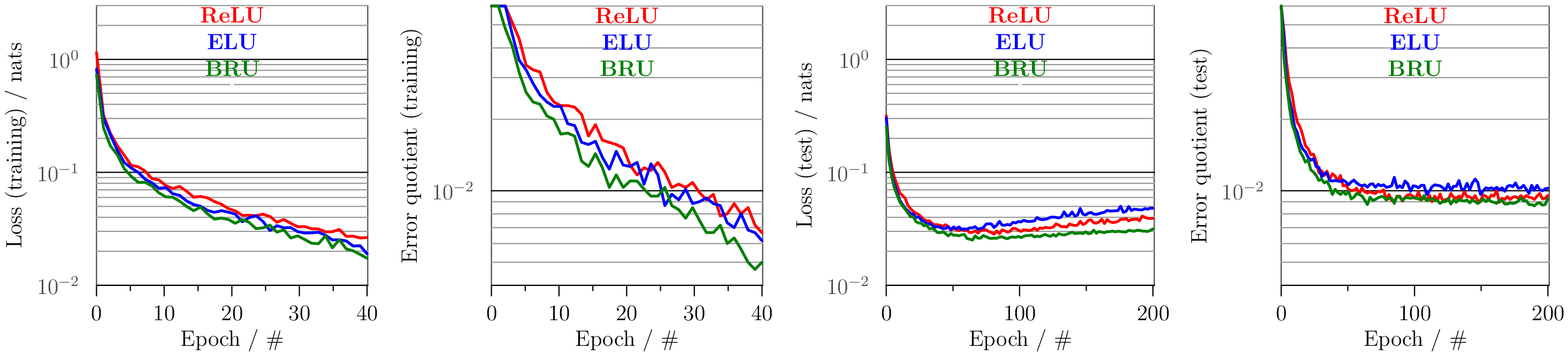} 

	\legend[fig:lenetmnist]{Simplified LeNet convolutional network training on MNIST data.}

\end{portrait}

\subsection{Convolutional networks with CIFAR data}

Simple network designs in the previous MNIST experiments were employed to moderate computational complexity for fair and
efficient comparison in performance between the activation functions. In order to benchmark more realistic models of
deep learning, a more complex network architecture was tested using CIFAR-10/100 data. The CIFAR-10/100 data sets each
comprise of 50000 training and 10000 test samples of 32x32x3 images with 10/100 label classifications. 

A 10 hidden-layer network design was based on a ConvPool architecture described previously
\citep{springenberg2014striving}, with minor modifications to combine same ('S') and valid ('V') padding for different
strides: [96, 3x3 convolution, 1S stride], [96, 2x2 convolution, 1V stride], [96, 3x3 max-pool, 2V stride], [192, 3x3
convolution, 1S stride], [192, 3x3 convolution, 1S stride], [192, 3x3 max-pool, 2V stride], [192, 3x3 convolution, 1S
stride], [192, 1x1 convolution, 1S stride], [10, 1x1 convolution, 1S stride], [10, 6x6 average pooling, 1S stride],
followed by a 10/100 fully-connected softmax output layer. For regularisation, dropout of 50\% was applied after each
max-pool layer. Images were zero-centered and scaled to unit variance (across all channels), and for each epoch training
images were bordered by single zero-value pixels and randomly cropped back to 32x32x3 sizes then horizontally flipped at
random. For bionodal networks, the following BRU transfer functions were assigned to the 10 hidden layers: [E3RU, O3RU,
E1RU, E2RU, O2RU, E1RU, E2RU, O2RU, E1RU, E1RU].

In Figure \ref{fig:cnncifar}, the loss and error quotients are plotted for each the three types of networks for the
training data (CIFAR-10 top, CIFAR-100 bottom) during the initial stage of learning and for the test data for all
epochs. The CIFAR-10 and CIFAR-100 training results again support the findings of \citet{clevert2015fast}, who show that
ELU convolutional networks train faster than ReLU networks. Consistent with the previous MNIST experiments, the bionodal
network exhibits the fastest learning. The results for the CIFAR-10 and CIFAR-100 data sets are notably similar. Both the
minimum test loss and test error quotient during training for the ELU networks are lower compared to that of ReLU
networks. However, the same measures show that bionodal networks outperform both ELU and ReLU networks. These results
confirm that with increases in data dimensionality and architectural complexity, the intrinsic capability of bionodal
networks to learn generalised models continues to exceed that of the other networks. 

\begin{portrait}[!htp] %[P] 
	
	\graphic[width=1.0\defwidth,trim={0 0 0 0},clip]{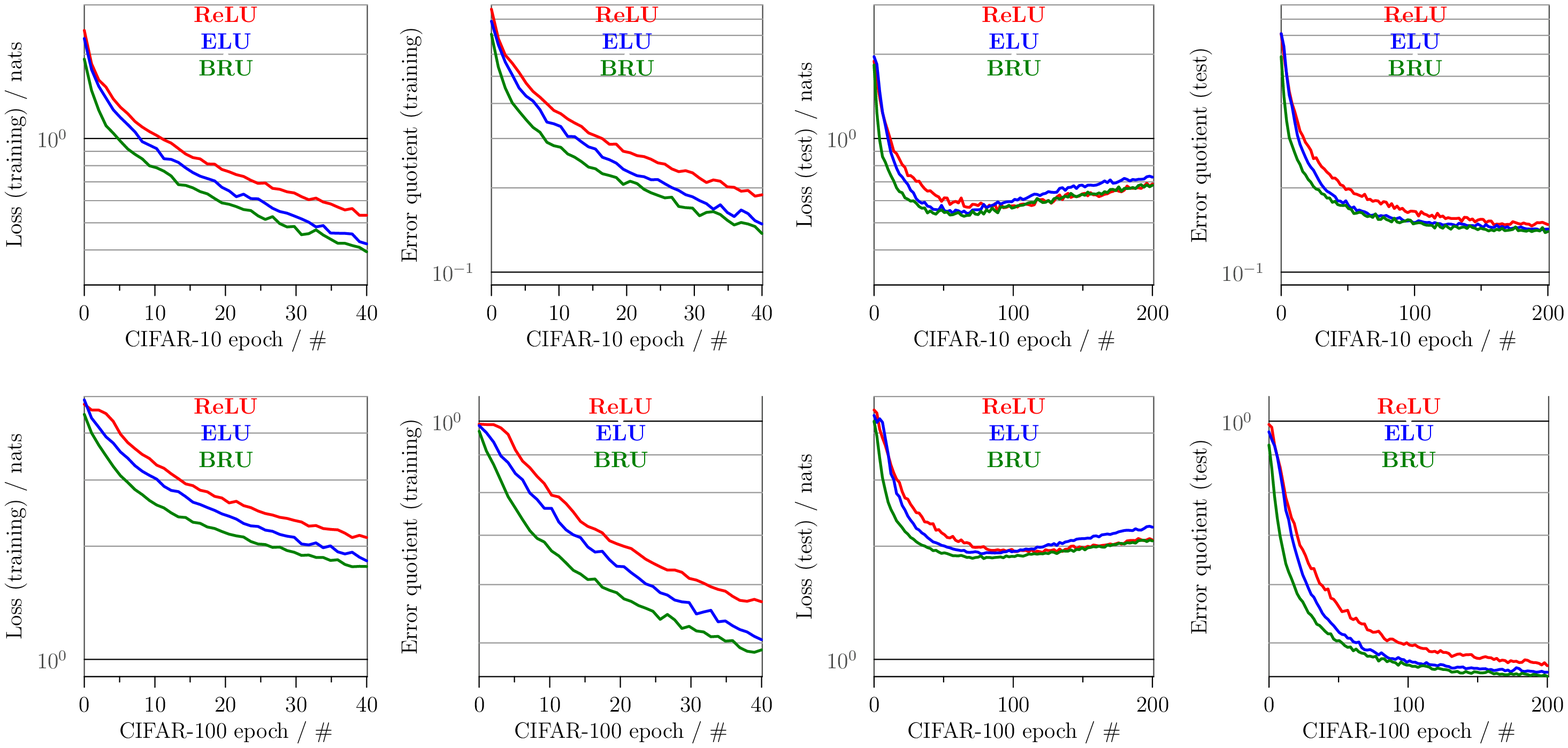} 

	\legend[fig:cnncifar]{ConvPool network training on CIFAR-10/100 data.}

\end{portrait}

\section{Discussion}

BRU activation functions were devised on basis of the biophysical properties of neuronal cells. While their derivation
was not a result of theoretical optimisations to deep learning models, the results of the present study suggest that
bionodal networks train more quickly and generalise better than their ReLU and ELU counterparts. Formal explanation
for these behaviours would clearly be merited for future work, but some intuitive deductions that might guide assignment
of BRU functions throughout the networks are considered.

\subsection{Speed}

For all hidden layers the learning speeds during ReLU and ELU training are intrinsically limited homogeneously by the
maximum of one in their activation gradients, whereas the maximum gradient for each bionodal layer corresponds to its
respective radix $r$.  Differences in the non-linearities across hidden layers in effect lead to different learning
speeds throughout the network. Assignment of higher radix BRU transfer functions to initial layers therefore results
naturally in faster learning compared to later layers in a manner that can be likened to early layer pre-training.
Gradient back-propagation from later layers however is most efficient with lower radix BRU transfer functions that
exhibit minimal saturation within their dynamic range.

The non-linearity of transfer functions in early layers is analogous to the response profile of sensory or afferent
components of the nervous system. Photoreceptors are one of few mammalian cell subtypes that do not exhibit discrete
all-or-none action potentials but graded action potentials. Response curves of photoreceptors to light exhibit
considerable non-linearity \citep{korenbrot2012speed}, conferring cones with the ability to respond to changes in light
intensity over several orders of magnitude. This natural phenomenon would suggest assignment of high radix BRU transfer
functions to early layers.

Transfer functions towards later layers however would be more analogous to the input-output relation of efferent
neuronal components. The final output cell type of the central nervous system is the motoneurone, which transmits
all-or-none action potentials to muscles to control movement. Progressive recruitment of motoneurones firing at
increasing frequencies induce greater forces of muscle contraction. It is notable that the input-output relation of
motoneurones is nearly linear \citep{heckman2009motoneuron}, consistent with the assignment of low radix BRU transfer
functions to later layers. 

\subsection{Generalisation}

Not only do bionodal networks learn more quickly, but also exhibit an intrinsic capacity to learn models that generalise
more effectively compared to ReLU and ELU networks. These findings might challenge the reported sparse encoding
advantages attributed to the ReLU transfer function \citep{glorot2011deep}. While the linearity in the ReLU and ELU
functions may alleviate the effects of vanishing gradients in backpropagation, it may come at the cost of compromising
sparsity in the positive activation encoding domain where an unchecked unit gradient risks an underlying tendency to
overfit.

It is therefore possible that the solution of mitigating the vanishing gradient using a na\"{i}ve linearity might be the
very cause of how resulting models are impeded in their generalisation without additional regularisation. A scaled
exponential linear unit transfer function has been introduced recently \citep{klambauer2017self} with reported
enhancement of regularisation in deep networks through self-normalisation. However this transfer function maintains a
linearity in the positive polarity and is therefore likely to be subject to the same impediment.

For radices greater than 1, both ERU and ORU subfamilies would tend towards more sparse encoding models compared to ReLU
and ELU transfer functions. Since ERU activations assume a more limited range of negative values, ERU layers would
inheritantly result in more sparse encoding than ORU layers of an equivalent radix. Inclusion of ORU transfer functions
within hidden layers however is useful because they bring activations within the inner hidden layers close to zero that
results in efficient natural gradient descent \cite{clevert2015fast}. While the use of symmetrical transfer functions
such as the hyperbolic tangent is known to improve deep learning convergence \citep{lecun1998efficient}, ORU
activations have the advantage of lacking asymptotic limits and therefore obviate the drawbacks of vanishing gradients.
The use of ORU functions may also introduce a natural form of self-normalisation with implicit regularisation.

\subsection{Future Work}

The aim of the present study was to obtain fair comparisons in performance of ReLU, ELU, and BRU transfer functions in
deep learning rather than submit a competitive attempt of improving upon current state-of-the-art benchmarks.
Optimisations based on weight regularisation, inception multi-convolution learning \citep{szegedy2015going},
batch-normalisation \citep{ioffe2015batch}, and residual learning \citep{he2016deep} have been instrumental in recent
deep learning advances. However most of these developments have been aimed at improving regularisation within ReLU
architectures. While future work would include assessment of how these optimisations could be exploited by bionodal
networks, it is likely that some adaption of the techniques will be necessary. For example L1 or L2 regularisation
techniques might be somewhat blunt instruments to apply homogeneously to bionodal networks since weight variance
initialisation is dependent on the radix which differs across layers.

While experiments have been performed using well established image data sets, it would instructive in future work to
extend the scope of research beyond computer vision since transfer functions are universal throughout deep learning
networks. For the purposes of simplicity, only very few activation functions from the ERU and ORU subfamilies were
used for the present study. It is highly likely that other permutation assignments throughout hidden layers using higher
radix non-linearities could improve results dramatically. Since the radix term $r$ is continuous, a potential learning
strategy would be to anneal non-linearities during the course of training to optimise sparsity during learning. While
empirical approaches were adopted for initialising the weight coefficients, it is probable that implementation of a full
theoretical framework underlying parameter initialisation for bionodal layers would further ameliorate learning
performance.

The suggestion that the logistic sigmoid function is `biological' would reflect a somewhat incomplete understanding of
the excitable properties of neuronal membranes. BRU transfer functions constitute activation non-linearities that are
substantially more physiologically plausible. In all experiments of the present study, bionodal networks outperformed
their ReLU and ELU counterparts both in terms of speed and generalisation often by considerable margins. It is perhaps
true therefore in the understanding of intelligent learning that we should not only look to biology for inspiration, but
for solutions.

\section{Acknowledgement}

The author thanks Maneesh Sahani for helpful discussion and comments on this work.

%\bibliography{bru}

\end{document}